\documentclass[runningheads]{llncs}

\usepackage{eccv}

\usepackage{eccvabbrv}

\usepackage{graphicx}
\usepackage{booktabs}
\usepackage{enumitem}
\usepackage{multirow}
\usepackage{xcolor}         
\usepackage{colortbl}
\usepackage{algorithm}
\usepackage{amsmath}
\usepackage{algorithmicx}
\usepackage[accsupp]{axessibility}  

\usepackage{hyperref}

\usepackage{orcidlink}

\definecolor{mygray}{gray}{0.9}

\begin{document}

\title{Target-aware Image Editing via Cycle-consistent Constraints} 

\titlerunning{FlowCycle}

\author{Yanghao Wang\and
Zhen Wang\and
Long Chen$^\dagger$}
\authorrunning{Y. Wang, Z. Wang, L. Chen}
\institute{The Hong Kong University of Science and Technology\\
\tt\small ywangtg@connect.ust.hk, \tt\small zhenwang@ust.hk, \tt\small longchen@ust.hk\\
\url{https://github.com/HKUST-LongGroup/FlowCycle}\\
}

\maketitle

\begin{figure*}[!h]
    \centering
    \includegraphics[width=0.95\linewidth]{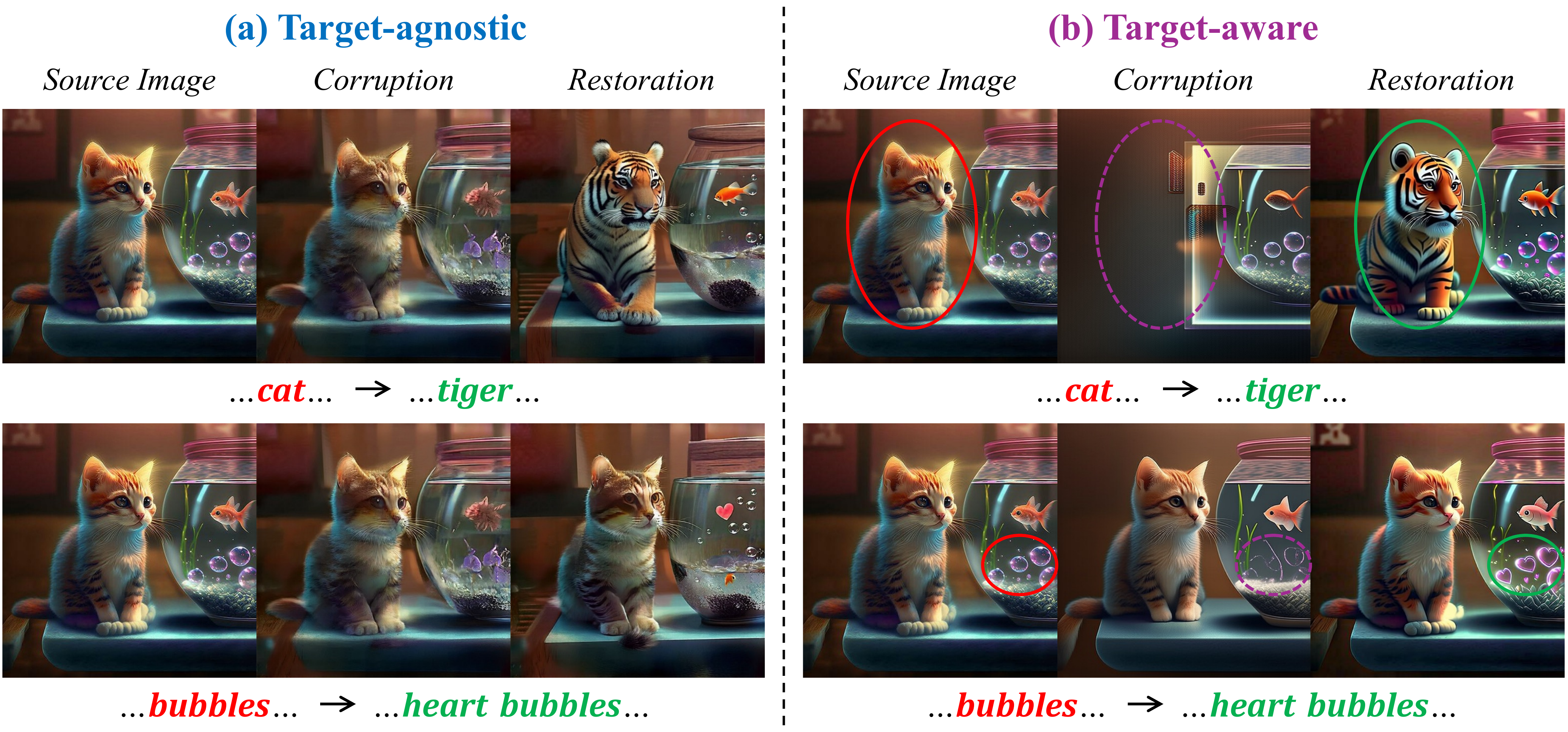}
    \vspace{-1em}
        \caption{\small{Comparisons between target-agnostic and target-aware corruption manner. 
}}
\vspace{-2em}
    \label{fig:1}
\end{figure*}
\begin{abstract}
Recent pre-trained text-to-image flow models have enabled remarkable progress in text-based image editing. Mainstream approaches adopt a \textbf{corruption-then-restoration} paradigm, where the source image is first corrupted into an editable ``intermediate state'' and then restored to the target image under the prompt guidance. However, current methods construct this intermediate state in a \textbf{target-agnostic} manner, \ie, they mainly focus on realizing source image reconstruction while neglecting the semantic gaps towards the specific editing target. This design inherently results in limited editability or inconsistency when the desired modifications substantially deviate from the source. In this paper, we argue that the intermediate state should be \textbf{target-aware}, \ie, selectively corrupting editing-relevant contents while preserving editing-irrelevant ones. Thus, we propose \textbf{FlowCycle}, an inversion-free and flow-based editing framework that parameterizes corruption with learnable noises and optimizes them through a cycle-consistent process. By iteratively editing the source to the target and recovering back to the source with dual consistency constraints, FlowCycle learns to produce a target-aware intermediate state, enabling faithful modifications while preserving source consistency. For efficiency, we further accelerate the optimization by dynamically adjusting the sampling steps. Extensive ablations demonstrated that FlowCycle achieves superior editing performance.
\end{abstract}

\begin{figure*}[!h]
    \centering
    \includegraphics[width=1\linewidth]{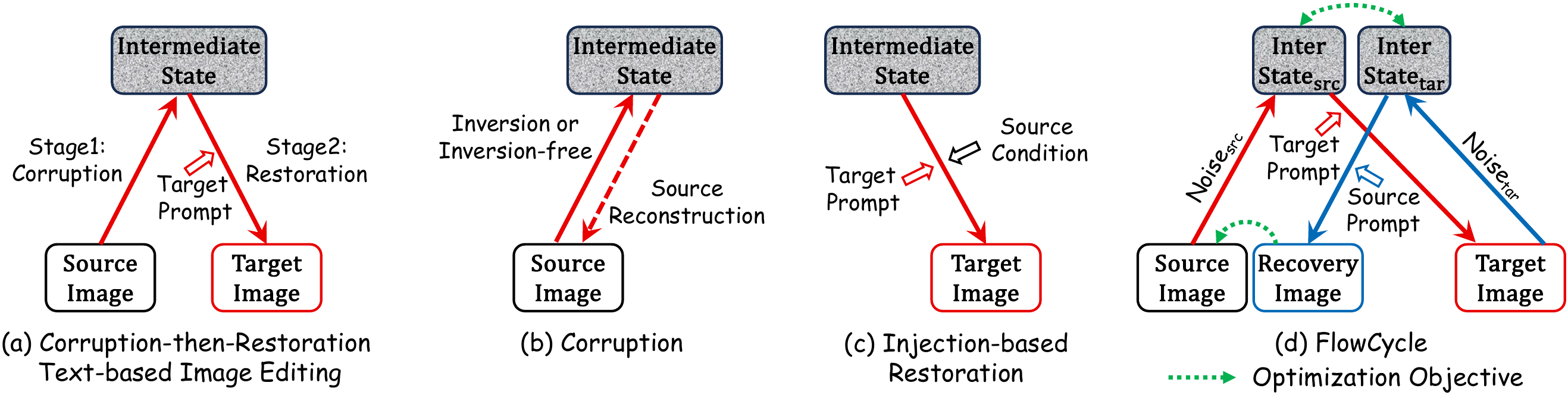}
        \caption{\textbf{Corruption-then-restoration paradigm for existing flow-based image editing methods}. Existing methods obtain target-agnostic intermediate states for subsequent editing. In contrast, FlowCycle obtains target-aware intermediate states via cycle-consistency optimization, which is both editable and source-preserving.}
    \vspace{-2em}
    \label{fig:2}
\end{figure*}
\section{Introduction}
\label{sec:1}

Pre-trained text-to-image flow models~\cite{esser2024scaling,flux2024} have recently proven strong generative capacity, which has gathered growing interest in adapting them for text-based editing~\cite{xu2024inversion,rout2024semantic,kulikov2024flowedit}. Unlike standard text-to-image generation, editing requires not only producing new content but also preserving certain source details. Specifically, given a source image and its description, the key challenge for text-based editing is to modify the image according to a target prompt while maintaining non-target content as the source, \ie, \emph{source consistency}. To achieve this goal, most existing flow-based editing approaches~\cite{xu2024inversion,rout2024semantic,kulikov2024flowedit,kim2025flowalign} closely follow well-established diffusion-based methods, since both flow-based and diffusion-based~\cite{ho2020denoising,song2020denoising} models share a similar generative principle, \ie, they corrupt the clean image with progressive noise and then learning to reverse this process to generate new samples. Consequently, existing flow-based editing methods generally adopt a common \emph{\textbf{corruption-then-restoration}} paradigm.
As shown in Figure~\ref{fig:2}(a), this paradigm consists of two main stages:
\begin{enumerate}[leftmargin=*]
    \item \textbf{Corruption.} The source image is first transformed into an \emph{intermediate state} (\ie, the corrupted image), which allows flexible editing while maintaining the source consistency required to restore the source image. As shown in Figure~\ref{fig:2}(b), the representative approaches include \emph{inversion-based} methods~\cite{deng2024fireflow,zhou2025multi} that perform ODE inversion~\cite{song2020denoising} and subsequent improvements~\cite{rout2024semantic,wang2024taming,kim2025reflex} to enhance the source reconstruction accuracy. As well as \emph{inversion-free} methods~\cite{kulikov2024flowedit,kim2025flowalign,mao2025tweezeedit} that directly sample noise and add it into the source image to improve the efficiency of the corruption process. 

    \item \textbf{Restoration.} Starting from the intermediate state, the pre-trained flow model restores the image under the guidance of the target prompt. To further enhance the preservation of source consistency, many methods inject source conditions (\cf, Figure~\ref{fig:2}(c)) during the restoration stage~\cite{wang2024taming,kim2025reflex,rout2024semantic,zhou2025multi}.
\end{enumerate}

In that case, an \emph{ideal intermediate state} of this paradigm should strike a balance between editability and preservation of source consistency. {As shown in Figure~\ref{fig:1}}(b), given a specific source-target prompt pair that aims to edit the cat (\eg, \texttt{$\dots$cat$\dots$} $\rightarrow$ \texttt{$\dots$tiger$\dots$}), the corruption should be applied to the editing-relevant contents (\eg, \texttt{cat}) to ensure faithful modifications, while the source information of editing-irrelevant contents (\eg, \texttt{fish tank}) should be preserved to maintain reliable source consistency. However, existing methods all generally acquire the intermediate state in a \emph{\textbf{target-agnostic}} manner, \ie, they emphasize the reconstruction of the entire source image without considering the specific editing targets, relying solely on the prior knowledge (\eg, source prompt and pre-trained model) to achieve this balance. {As shown in Figure~\ref{fig:1}(a)}, such intermediate states reveal an approximate reconstruction of the source image, preserving unbiased global layout structures and semantic details. When restored with different target prompts focusing on distinct contents, this source-aware but target-agnostic design often leads to unsatisfactory results, especially when the editing target significantly deviates from the source image or the prior knowledge. For instance, as shown in the Figure~\ref{fig:1}(a), when the target is to change the shape of the bubble (\eg, \texttt{$\dots$bubbles$\dots$} $\rightarrow$ \texttt{$\dots$heart bubbles$\dots$}), the target-agnostic intermediate state not only fails to reliably preserve editing-irrelevant contents (\eg, \texttt{cat} and \texttt{goldfish}) but also conflict with editing-relevant contents (\eg, \texttt{bubbles}), making them difficult to modify as desired. Meanwhile, even injecting additional source conditions during restoration cannot resolve the problem, since the intermediate state itself remains misaligned with the editing target.

In this paper, we argue that the \emph{ideal intermediate state} should be constructed in a \emph{\textbf{target-aware}} manner, selectively corrupting editing-relevant contents while preserving editing-irrelevant ones based on specific editing targets to enable faithful and effective editing. 
To this end, we introduce \textbf{FlowCycle}, a novel inversion-free and flow-based framework that faithfully incorporates target awareness, enabling precise and reliable text-based image editing. Notably, such a target-aware state exhibits a natural symmetry between the source and target image: the editing-relevant regions should be interchangeable, while the editing-irrelevant regions should remain invariant across both. Motivated by this, we construct a \emph{cycle} between the source and target image, allowing the model to exploit this inherent symmetry and learn a target-aware corruption.

Specifically, instead of relying on fixed inversion or random noise sampling, our method parameterizes the corruption process with learnable noises through a cycled framework. As shown in Figure~\ref{fig:2}(d), it consists of three main steps. \emph{1) Source to target} (\ie, the red path of Figure~\ref{fig:2}(d)): the source image is first corrupted with a learnable source noise and then restored under the guidance of the target prompt, producing an initial target image that reflects the desired edit.
\emph{2) Target to source} (\ie, the blue path of Figure~\ref{fig:2}(d)): the initial target image is also corrupted with another learnable target noise and then restored with the source prompt, yielding a recovery image that traces back to the source. \emph{3) Cycle-consistent Optimization}: we impose two constraints to optimize the learnable noises — (i) align the recovery image to the source image, and (ii) align the corrupted source image with the corrupted target image. This joint regularization stabilizes the optimization and enables target-aware corruption. By iterating this process, learned noises gradually produce a target-aware intermediate state (\cf, Figure~\ref{fig:1}(b)), where the corrupted source image and the corrupted target image share an ideal role, \ie, corrupt the editing-relevant content while maintaining the editing-irrelevant parts based on a specific editing target. 

Although this optimization process introduces additional computational overhead due to iterative generation, it can be naturally accelerated by leveraging the sampling flexibility of flow models. Accordingly, we further propose a progressive sampling strategy that dynamically adjusts the number of sampling steps across optimization rounds to improve efficiency while maintaining editing quality.


In summary, we made three contributions in this paper:
\begin{itemize}
    \item We identify a fundamental limitation in prevalent corruption-then-restoration editing methods: the intermediate state is acquired in a \emph{target-agnostic} manner, which often leads to unsatisfactory results. Moreover, we argue that the target-aware paradigm is of significant necessity.
    \item To achieve target-aware editing, we propose the first \emph{target-aware} corruption strategy and introduce FlowCycle, an inversion-free, flow-based framework that adaptively focuses corruption on editing-relevant content while preserving irrelevant content. Meanwhile, we introduce an efficient acceleration strategy to reduce the computational overhead of the optimization process.
    \item Extensive experiments across benchmarks and diffusion backbones have empirically shown the robust effectiveness of FlowCycle over existing methods. 
\end{itemize}

\section{Related Work}
\label{sec:2}

Existing text-based image editing methods largely converge on the \emph{corruption-then-restoration} paradigm, improving editing performance at either the corruption or the restoration stage. While a few alternative approaches, such as unified diffusion models~\cite{fu2025univg,xiao2025omnigen,wu2025qwen,wang2025selftok} or MLLM-driven models~\cite{deng2025emerging,ai2025ming}, also contributed to the field, this paper focuses primarily on methods within this framework due to its broad adoption and foundational role.

\noindent\textbf{Inversion-based Corruption}. Methods in this category construct the intermediate state from the source image via ODE inversion~\cite{song2020denoising,deng2024fireflow,zhou2025multi}, which allows the preservation of structural and semantic details for reconstruction (\ie, restore with an empty prompt). However, the iterative inversion is prone to error accumulation~\cite{huberman2024edit,mokady2023null}, which degrades the fidelity of source information and weakens the reliability of subsequent editing. Thus, recent works focus on improving inversion accuracy~\cite{rout2024semantic,wang2024taming,kim2025reflex}, enabling more faithful reconstruction of the source image and providing a stronger foundation for high-quality editing.

\noindent\textbf{Inversion-free Corruption}. To alleviate the high computational overhead caused by ODE inversion, inversion-free methods~\cite{kulikov2024flowedit,kim2025flowalign,mao2025tweezeedit} bypass the inversion step by directly sampling random noise to the source image to construct the intermediate state. Compared with inversion-based strategies, these methods achieve significantly faster corruption. However, they remain highly sensitive to noise level~\cite{meng2021sdedit} and hyperparameter choices~\cite{kulikov2024flowedit,kim2025flowalign}. Excessive corruption risks erasing key source information, while insufficient perturbation limits editability. Thus, the central challenge lies in striking the right balance between editability and faithful preservation of source consistency. 

\noindent\textbf{Injection-based Restoration}. Solely relying on the intermediate state for restoration often results in partial information loss and inconsistency in the edited image~\cite{huberman2024edit}. To address this, a number of methods propose to explicitly inject source information throughout the restoration process, ensuring stronger alignment between the edited image and the source image. This can be achieved through additional model inputs, \eg, optimized prompt~\cite{mokady2023null} and latent~\cite{wu2023latent,huberman2024edit}. Intermediate model features~\cite{wang2024taming,kim2025reflex} derived from the source image, \eg, attention maps~\cite{hertz2022prompt, tumanyan2023plug}. And enhanced model outputs~\cite{rout2024semantic,zhou2025multi} such as predicted noise or velocity. Nevertheless, without a properly constructed and target-aware intermediate state, such injection strategies still offer limited improvements.

\noindent\textbf{Cycle Consistency in Generation}. Well-recognized cycle consistency~\cite{zhu2017unpaired} has also been explored, particularly in image-to-image translation tasks, where consistency regularization is enforced through translation cycles between source and target domains. Such methods typically require images from two specific domains for training or optimization~\cite{sasaki2021unit,xu2023cyclenet,leetext}. While recent work has attempted to adapt cycle consistency to image editing~\cite{beletskii2025inverse}, their ``cycle" still focuses solely on source inversion and reconstruction. In contrast, our work leverages cycle consistency to enable a novel \emph{target-aware} corruption during text-based editing, where the cycle is established between source and target to explicitly align editing-relevant and irrelevant contents.

\section{Method}
\label{sec:3}

\subsection{Preliminaries: Flow Matching (FM)}
\label{sec:3.1}
Flow Matching (FM) models~\cite{lipman2022flow,liu2022flow} are trained to fit a velocity field $u_t(x_t)$ on time $t \in [0,1]$ such that they can transport data from one distribution (\eg, normal distribution $\pi_1$) to another distribution (\eg, images distribution $\pi_0$). Specifically, given a data point $x_1 \sim \pi_1$, the transport path can be simulated by solving an ordinary differential equation (ODE):
\begin{equation}
    \label{eq:1}
    dx_t = u_t(x_t)dt,
\end{equation}
which converts $x_1$ into $x_0 \sim \pi_0$. To match the underlying velocity field $u_t(x_t)$, a velocity prediction network~\cite{chen2018neural} $v_\theta(x_t,t)$ was trained by minimizing the such loss:
\begin{equation}
    \label{eq:2}
    \mathcal{L}_{FM} = \mathbb{E}_{x_t,t}\|v_\theta(x_t,t) - u_t(x_t)\|^{2}.
\end{equation}
However, the ground-truth velocity field $u_t(x_t)$ is not accessible. Alternatively, prior studies~\cite{lipman2022flow,liu2022flow} propose to optimize the approximated conditional FM loss:
\begin{equation}
    \label{eq:3}
    \mathcal{L}_{CFM} = \mathbb{E}_{x_0,p_t(x_t|x_0),t}\|v_\theta(x_t,t) - u_t(x_t|x_0)\|^{2},
\end{equation}
where $p_t(x_t|x_0)$ is usually defined as the optimal transport~\cite{mccann1997convexity} probability path, \ie, $p_t(x_t|x_0) = \mathcal{N}(x_t;(1-t)x_0,t^2I)$ while $I$ is the identity matrix. Since $\pi_1$ is standard normal distribution, we can sample $x_1 \sim \pi_1$ and get:
\begin{equation}
    \label{eq:4}
    x_t = (1-t)x_0 + tx_1 \Rightarrow u_t(x_t|x_0) = \frac{dx_t}{dt} = x_1 - x_0.
\end{equation}
By substituting $u_t(x_t|x_0) = x_1 - x_0$ into Eq.~\eqref{eq:3}, get final training objective:
\begin{equation}
    \label{eq:5}
    \mathcal{L}_{CFM} = \mathbb{E}_{x_0\sim\pi_0,x_1\sim\pi_1,t}\|v_\theta(x_t,t) - (x_1 - x_0)\|^{2}.
\end{equation}
In the inference stage, we can start from a randomly sampled Gaussian noise $x_1 \sim \pi_1$ and then integrate to generate a clean image: $x_0 = x_1 -\int_1^0 v_\theta(x_t,t)dt$.

\subsection{Corruption-then-Restoration Paradigm}
\label{sec:3.2}
Given a source image $x_0^{src}$, source prompt $c_{src}$, and target prompt $c_{tar}$, the paradigm has two stages: 

\noindent 1) \underline{Corruption}: Convert the source image $x_0^{src}$ into an intermediate state $x_t^{src}$ 
:
\begin{equation}
    \label{eq:6}
    x_t^{src} = \psi(x_0^{src},t, c_{src}), \quad t \in (0,1],
\end{equation}
where $\psi$ is a corruption function such as directly adding random noise (inversion-free) or calculating an inversion (inversion-based). The principle of both corruptions is obtaining an intermediate state $x_t^{src}$ while maintaining the information of the source image $x_0^{src}$. In that case, starting from $x_t^{src}$ can benefit \textit{source consistency}, \ie, the editing result can maintain the editing-irrelevant part.
It is worth noting that current corruption is target-agnostic, \ie, the noisy image $x_t^{src}$ is fixed for different editing targets. In contrast, we aim to find an ideal target-aware $x_t^{src}$ that can specifically destroy the editing-relevant part while maintaining the editing-irrelevant part.

\noindent 2) \underline{Restoration}: After getting $x_t^{src}$, the most straightforward approach is to take it as the starting point of restoration directly (\ie, $x_t^{tar} = x_t^{src}$) and denoising it under the guidance of the target prompt $c_{tar}$. However, only relying on the starting point that contains the source image's information is not enough for \textit{source consistency}. Thus, existing methods inject a drift item $\Delta_t$ (containing the source image's information) during the denoising:
\begin{equation}
    \label{eq:7}
    dx_t^{tar} = \xi(v_\theta,x_t^{tar},t,c_{tar},\Delta_t)dt,
\end{equation}
where $\xi$ represents different injection functions. For instance, adding $\Delta_t$ to the middle attention maps, predicted velocity $v_\theta$, or input $x_t^{tar}$. The principle is enhancing the \textit{source consistency} of the final editing result by injecting a source-relevant drift item $\Delta_t$ during denoising. This drift item can rectify the denoising path and help the final editing result maintain more editing-irrelevant parts.

Our motivation is to find a smart target-aware intermediate state $x_t^{src}$ as the start point of restoration. In that case, even if there is no source condition injection (\ie, $\Delta_t$), we can still gain a good editing result since the source information is preserved in the intermediate state quite well. In the following Section~\ref{sec:3.3}, we propose one optimization-based solution to find this ideal $x_t^{src}$.
\begin{figure*}[!t]
    \centering
    \includegraphics[width=1\linewidth]{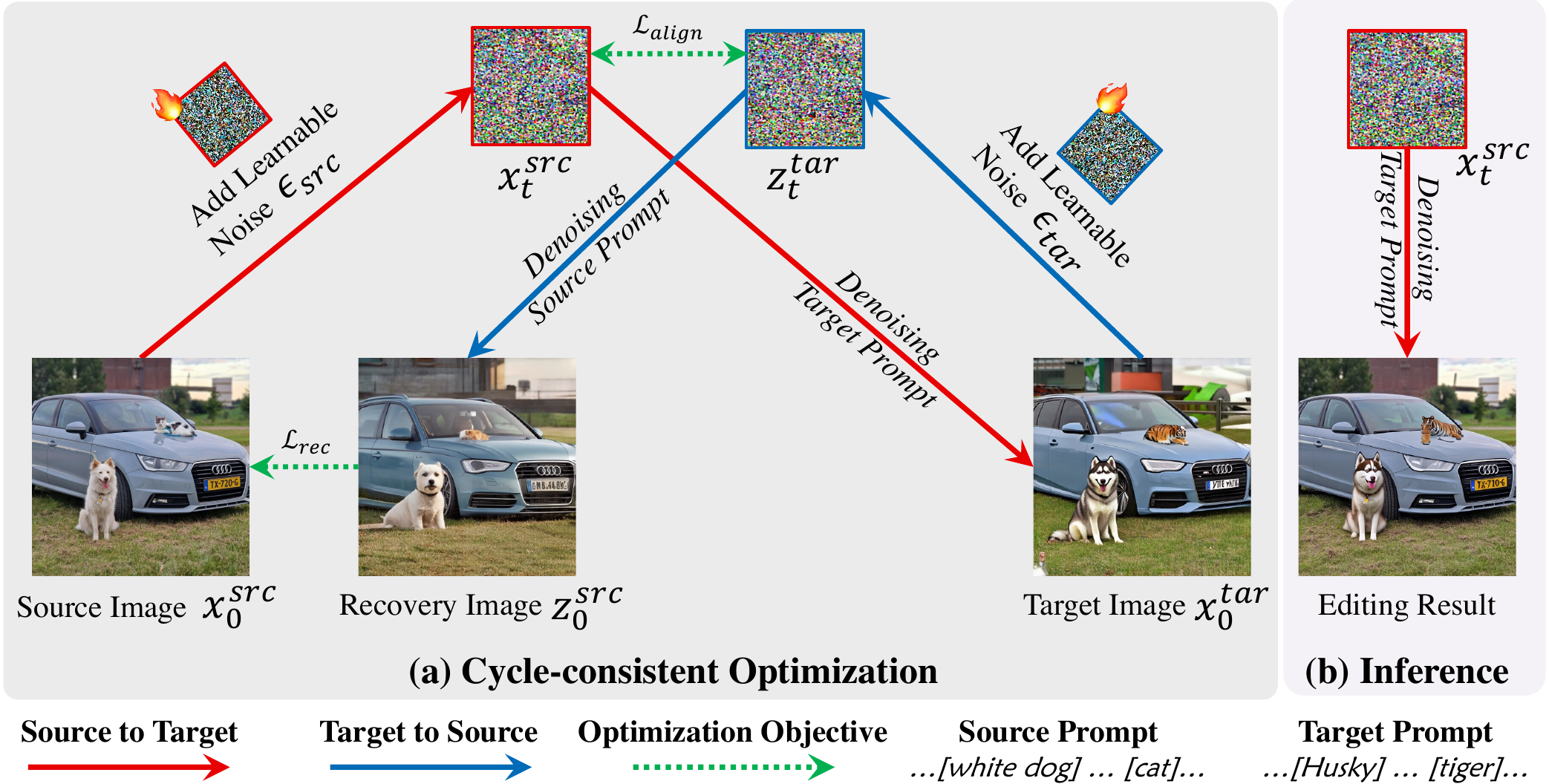}
        \caption{\textbf{Framework of FlowCycle}. We rectify the corruption process by optimizing two learnable noises (\ie, $\epsilon_{src}$ and $\epsilon_{tar}$) under the cycle-consistent constraints of whole editing. After optimization, a target-aware intermediate state $x_t^{src}$ can be obtained by adding $\epsilon_{src}$ to the source image $x_0^{src}$. Starting the restoration from this optimized $x_t^{src}$, the quality of the editing result $x_0^{tar}$ can be significantly improved.}
    \label{fig:3}
    \vspace{-1em}
\end{figure*}
\subsection{FlowCycle}
\label{sec:3.3}

\textbf{Overview.} As shown in Figure~\ref{fig:3}(a), our FlowCycle contains three steps, including:
1) \underline{Source to Target}: We first add a parameterized learnable noise $\epsilon_{src}$ (initialized as a Gaussian noise) to the source image $x_0^{src}$ and get the intermediate state $x_t^{src}$. After getting $x_t^{src}$, we leverage the pretrained flow matching model to denoise it conditioned on the target prompt $c_{tar}$. The denoised result is the target image $x_0^{tar}$.
2) \underline{Target to Source}: After getting the initialized target image $x_0^{tar}$, we add another parameterized learnable noise $\epsilon_{tar}$ (initialized as a Gaussian noise) to it and then get the noisy image $z_t^{tar}$. After that, we use the flow matching model conditioned on the source prompt $c_{src}$ to denoise it into a recovery image $z_0^{src}$.
3) \underline{Cycle-consistent Optimization}: After (1) and (2), we calculate Mean Squared Error (MSE) loss between $x_t^{src}$ and $z_t^{tar}$ as well as the MSE loss between $x_0^{src}$ and $z_0^{src}$. Then we optimize the parameterized $\epsilon_{src}$ and $\epsilon_{tar}$ to decrease the loss.
By repeating the three steps, $\epsilon_{src}$ and $\epsilon_{tar}$ are iteratively optimized. 
In the inference stage (\cf, Figure~\ref{fig:3}(b)), the optimized $\epsilon_{src}$ is used for getting the target-aware intermediate state $x_t^{src}$. Then we can directly denoise from this intermediate state under the guidance of the target prompt without any other source condition injection and gain a satisfactory editing result.

\subsubsection{Source to Target.}
\label{sec:3.3.1}
We first randomly sample a noise $\epsilon_{src}$ from the standard normal distribution and parameterize it as a learnable vector. Then, we follow the forward process of FM to add $\epsilon_{src}$ to the source image $x_0^{src}$:
\begin{equation}
    \label{eq:8}
    x_t^{src} = (1-t) * x_0^{src} + t * \epsilon_{src} \quad t\in (0,1),
\end{equation}
where $x_t^{src}$ is the intermediate state obtained by linearly interpolating the source image and noise. The interpolation ratio is controlled by $t$. We take $x_t^{src}$ as the starting point of the following restoration process, \ie, $x_t^{tar} = x_t^{src}$. By inputting $x_t^{tar}$ into the pretrained text-to-image FM model (\ie, the velocity prediction network $v_\theta$) under the guidance of target prompt $c_{tar}$, we can get the predicted velocity $v_\theta(x_t^{tar},t,c_{tar})$, which can transport $x_t^{tar}$ to $x_0^{tar}$:
\begin{equation}
    \label{eq:9}
    x^{tar}_{t-\Delta} = x^{tar}_{t} - v_\theta(x_t^{tar},t,c_{tar}) * \Delta \Rightarrow x_0^{tar},
\end{equation}
where $\Delta$ is the step size. After the above process, we can get a naive target editing image $x_0^{tar}$ (\eg, SDEdit~\cite{meng2021sdedit} directly takes it as the final editing result). However, it turns out $x_0^{tar}$ suffers from the low \textit{source consistency} because the random corruption indiscriminately destroyed the information of the source image.

\subsubsection{Target to Source.}
\label{sec:3.3.2}
To further improve $x_0^{tar}$, we try to recover the source image from $x_0^{tar}$. Considering its symmetrical nature, a better recovery result indicates that $x_0^{tar}$ is of higher \textit{source consistency}. Specifically, we parameterize another noise $\epsilon_{tar}$ (initialized from standard Gaussian noise) and add it to $x_0^{tar}$:
\begin{equation}
    \label{eq:10}
    z_t^{tar} = (1-t) * x_0^{tar} + t * \epsilon_{tar} \quad t\in (0,1),
\end{equation}
where $z_t^{tar}$ is the noised version (the intermediate state of the reverse editing process) of the generated target image $x_0^{tar}$. We take $z_t^{tar}$ as the starting point of the following restoration process, \ie, $z_t^{src} = z_t^{tar}$ and denoise it under the guidance of source prompt $c_{src}$:
\begin{equation}
    \label{eq:11}
    z^{src}_{t-\Delta} = z^{src}_{t} - v_\theta(z^{src}_{t},t,c_{src}) * \Delta \Rightarrow z^{src}_{0},
\end{equation}
where $z^{src}_{0}$ is the recovery of source image. The better recovery (\ie, $z^{src}_{0}$ is closer to $x^{src}_{0}$) indicates that the generated target image $x^{tar}_{0}$ maintains editing-irrelevant information of the source image better. After the above two steps, we can conduct the following cycle-consistent optimization.

\subsubsection{Cycle-consistent Optimization.}
\label{sec:3.3.3}
As shown in Figure~\ref{fig:3}(a), the recovery image $z^{src}_0$ is quite different from the source image $x^{src}_0$ because the two intermediate states (\ie, $x_t^{src}$ and $z_t^{tar}$) of the whole cycle were indiscriminately destroyed and the global source information was lost. To get the ideal target-aware intermediate state, we constrain the whole cycle to satisfy two cycle-consistent objectives:

\noindent1) The two intermediate states $x_t^{src}$ and $z_t^{tar}$ should share the aligned semantics as corrupted images, \ie, destroy the editing-relevant parts of the source image (\eg, \texttt{white dog} and \texttt{cat}) while maintaining the editing-irrelevant parts (\eg, \texttt{background} and \texttt{car}). To achieve that, we use an alignment constraint to force $x_t^{src}$ and $z_t^{tar}$ to be close in the latent space:
\begin{equation}
    \label{eq:12}
    \mathcal{L}_{align} = \|x_t^{src} - z_t^{tar}\|^2_2.
\end{equation}
2) Since $x_t^{src}$ and $z_t^{tar}$ share the same semantics, starting from $z_t^{tar}$, the recovery result (\ie, $z_0^{src}$) should be highly similar to the source image $x_0^{src}$ because the ideal $z_t^{tar}$ can maintains the editing-irrelevant parts. Thus, we take a recovery constraint to encourage $z_t^{tar}$ to be a good start point of recovering to the source image $x_0^{src}$:
\begin{equation}
    \label{eq:13}
    \mathcal{L}_{rec} = \|z_0^{src} - x_0^{src}\|^2_2.
\end{equation}
To decrease the above two losses, we leverage the property that $\epsilon_{src}, \epsilon_{tar}$ can effectively rectify $x_t^{src}$ and $z_t^{tar}$ based on Eq.~\eqref{eq:8} and Eq.~\eqref{eq:10} and further rectify the whole cycle. Thus, we can jointly optimize the two learnable noises to decrease the $\mathcal{L}_{align}$ and $\mathcal{L}_{rec}$:
\begin{equation}
    \label{eq:14}
    \epsilon_{src}, \epsilon_{tar} = \operatorname*{argmin}_{\epsilon_{src}, \epsilon_{tar}}( \mathcal{L}_{rec} + \lambda \mathcal{L}_{align}),
\end{equation}
where $\lambda$ is a weight hyperparameter to balance the two loss items. Through the above cycle-consistent optimization, $\epsilon_{src}, \epsilon_{tar}$ can selectively destroy the information according to whether it is editing-relevant. Then the intermediate state $x_t^{src}$ can be rectified towards the ideal role to be both editable and editing-irrelevant preserved. During inference, we can directly denoise the target-aware intermediate state $x_t^{src}$ to get an improved editing result (\cf, Figure~\ref{fig:3}(b)).

\subsection{Optimization Acceleration}
\label{sec:3.4} 

Our primary motivation is to obtain a target-aware intermediate state that preserves editing-irrelevant regions while making editing-relevant parts mutable. As the first work in this direction, FlowCycle offers a straightforward and effective solution. Although this optimization-based framework inevitably introduces higher computational cost than optimization-free alternatives, its efficiency can be naturally improved by exploiting the inherent sampling flexibility of flow models. Specifically, the dominant computational overhead arises from generating two images (target image $x_0^{tar}$ and recovery image $z_0^{src}$) within each optimization round, resulting in a large number of function evaluations (NFEs). Therefore, accelerating the sampling trajectory during generation provides a direct way to reduce the overall cost. This can be achieved through dynamic step scheduling, advanced ODE solvers, or early-stopping strategies, all of which are compatible with our framework.

In this paper, we pioneer a simple yet effective acceleration strategy by dynamically adjusting the sampling steps during optimization. The key observation is that in early optimization rounds, the learnable noise is far from the ideal state, thus we can use fewer sampling steps. As the optimization proceeds, more fine-grained updates are required, motivating the use of increased sampling steps in later rounds. Therefore, we a design monotonically increasing sampling schedule to decide the sampling steps for each optimization round. For example, the scheduler can be:
\begin{equation}
    \label{eq:15}
    f(n) = \text{rounding}(\sin(\frac{\pi n}{2N})*M),
\end{equation}
where $N$ is the overall optimization rounds, $n$ is the current optimization round index, and $M$ is the largest sampling steps. Moreover, this scheduler can be flexibly replaced by other monotonically increasing functions, such as linear or exponential schedules, to further enhance the acceleration effect.

\section{Experiments}
\label{sec:4}
\noindent\textbf{Implementation Details}. We used SD-3-medium~\cite{esser2024scaling} from HuggingFace as the pretrained text-to-image flow model. The overall timestep number is set to $50$ and the corruption step $t=33$ (\cf, Eq.~\eqref{eq:8}). We used the Adam optimizer~\cite{kingma2014adam} with $1e-1$ learning rate for 100 rounds of noise optimization. The loss weight $\lambda$ was set to $0.2$. More details are in the Appendix and the codes.

\begin{figure*}[!t]
    \centering
    \includegraphics[width=0.8\linewidth]{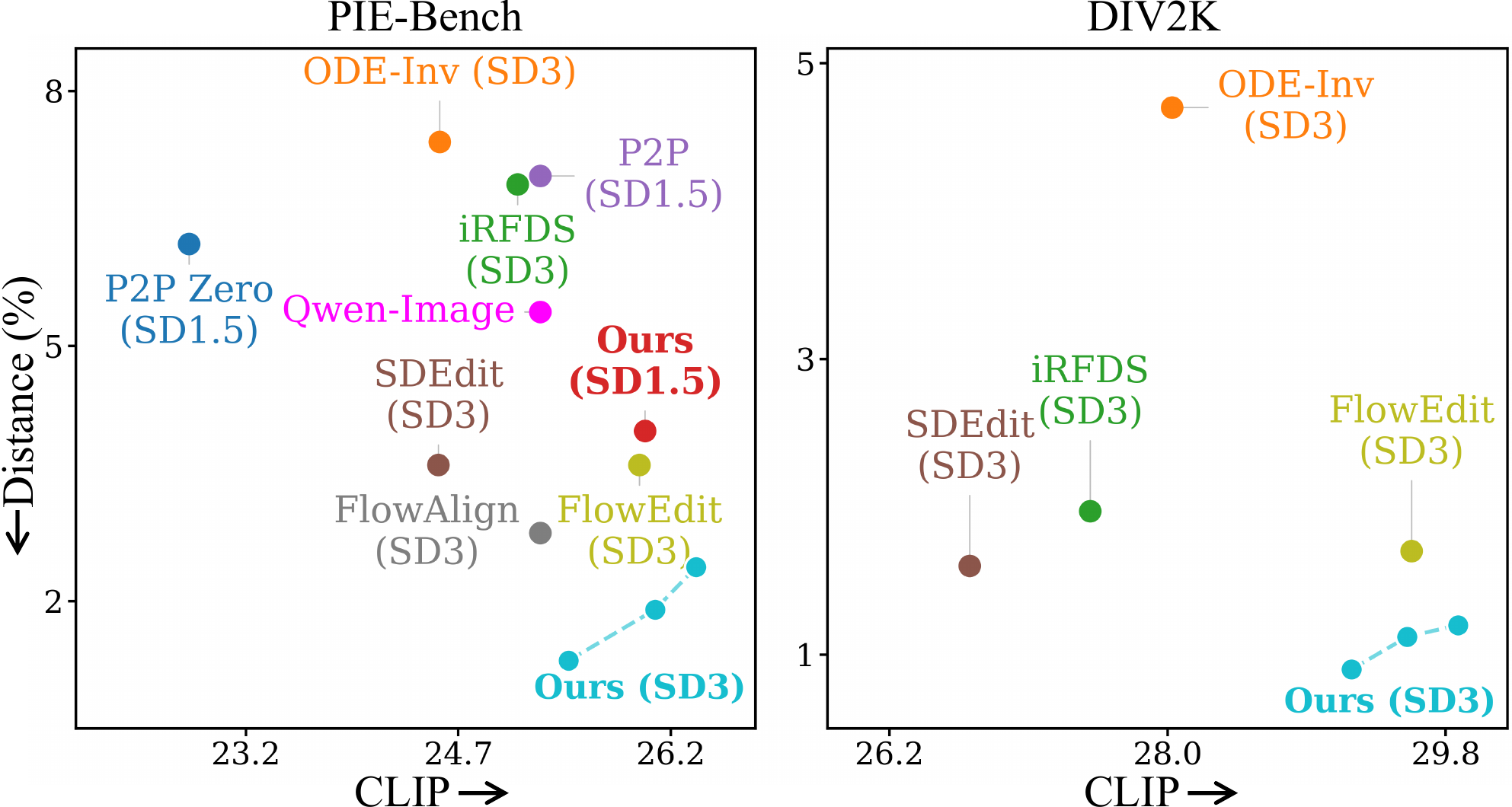}
        \caption{\textbf{Quantitative comparisons}. The closer to the bottom right corner, the better the balance between Structure Distance (source consistency) and CLIP Score (semantic alignment). Connected markers represent different hyperparameters (see Appendix). Ours can achieve the best balance across datasets and diffusion models.}
    \label{fig:4}
\end{figure*}

\noindent\textbf{Benchmarks}. We evaluated our method on two benchmarks. 1) PIE-Bench~\cite{ju2023direct}: a text-based image editing benchmark. It contains 700 (source image, source prompt, target prompt) tuples, which cover nine editing types (\eg, ``add object'' and ``change style''). Each image is $512^2$ resolution. Besides, there is usually only a single editing target in each prompt. 2) DIV2K~\cite{agustsson2017ntire} subset: containing over 70 real images with around 280 (source image, source prompt, target prompt) tuples. Each image is $1024^2$ resolution. Meanwhile, this dataset has longer target prompts that may contain multiple editing targets, which is more challenging.
For the evaluation metrics, we reported the \textit{CLIP Score}~\cite{hessel2021clipscore} (for semantic alignment evaluation) and \textit{Structure Distance}~\cite{tumanyan2022splicing} (for source consistency evaluation). More details about benchmarks are in the Appendix.
\begin{figure*}[!t]
    \centering
    \includegraphics[width=1\linewidth]{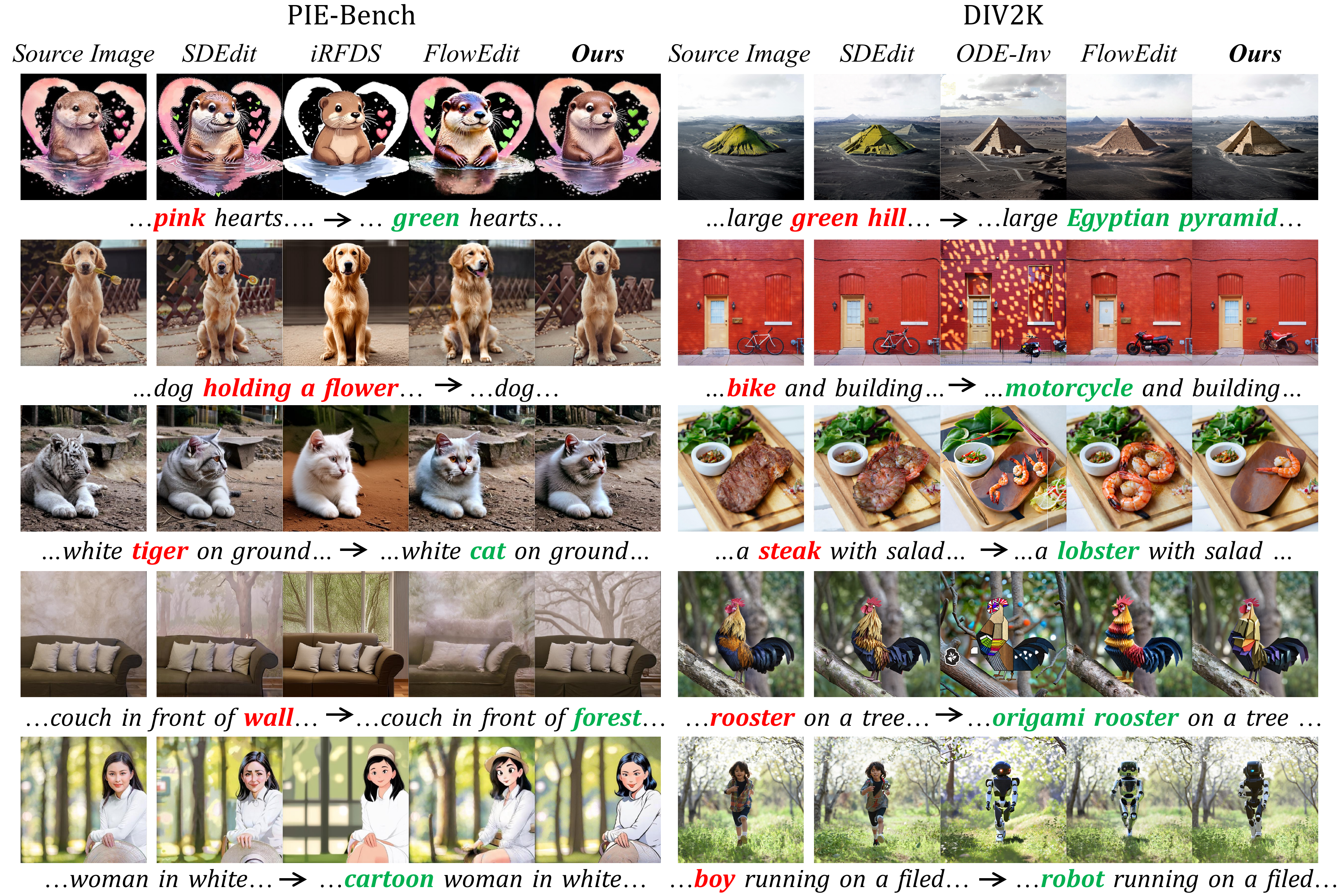}
        \caption{\textbf{Qualitative comparisons on $512^2$ resolution PIE-Bench (left half) and $1024^2$ resolution  DIV2K (right half)}. Compared with various baselines across different editing types (\eg, object or style change), FlowCycle can significantly maintain the editing-irrelevant parts of the source image while adhering to the editing target. Meanwhile, these two datasets cover different resolutions and editing complexity (\eg, DIV2K contains multi-object edit cases), indicating our robust effectiveness.}
    \label{fig:5}
\end{figure*}

\subsection{Comparison with State-of-the-Art Methods}
\label{sec:4.1}

\noindent\textbf{Baselines}. We compared our method with five state-of-the-art flow-based editing methods that were implemented with SD-3-medium: 1) \textit{ODE inversion} (\textit{ODE-Inv}), which gets the inversion by using the Euler Solver and denoises with the target prompt. 2) \textit{SDEdit}~\cite{meng2021sdedit}, which adds random noise and then denoises with the target prompt. 3) \textit{iRFDS}~\cite{yang2024text}, which is a SDS-based editing method. 4) \textit{FlowEdit}~\cite{kulikov2024flowedit} and 5) \textit{FlowAlign}~\cite{kim2025flowalign} are two inversion-free editing methods. The results and hyperparameters for the reimplementation of these methods follow previous works. Moreover, since the target-aware editing manner is not limited to Flow-based models, our method can also be compatible with other diffusion backbones like DDPM-based models. Thus, we also compared our method with Prompt-to-Prompt~\cite{hertz2022prompt} (\textit{P2P}) and pix2pix-zero~\cite{parmar2023zero} (\textit{P2P Zero}) based on SD-1.5~\cite{rombach2022high}. We also compared with the editing model Qwen-Image-Edit-2509 (\textit{Qwen-Image})~\cite{wu2025qwen}.
More details are in the Appendix.

\noindent\textbf{Qualitative Evaluation}. We gave qualitative comparisons in Figure~\ref{fig:5}. We have three observations: 1) Our method produced high-quality editing results (high source consistency and semantic alignment) on cases from various editing types. This indicates that our method is generally effective for extensive editing requirements. 2) Highlightly, our method can significantly maintain the editing-irrelevant parts. This advantage exactly verifies that the target-aware corruption can smartly hold the editing-irrelevant source content while making the editing-relevant contents editable. 3) Across two different datasets (different resolution and prompt complexity), FlowCycle demonstrates consistent superiority, indicating our robust effectiveness. More qualitative results are in the Appendix.
\noindent\textbf{Quantitative Evaluation}. As shown in Figure~\ref{fig:4}, by only optimizing noises under the cycle constraints and getting a target-aware intermediate state, our method can maintain a remarkable source consistency (\ie, the lower \textit{Structure Distance}). The source consistency is a significant outcome benefited from the cycle consistency constraints. At the same time, our method can also achieve a good semantic alignment (\ie, the higher \textit{CLIP score}) to the target prompt. Overall, our method can achieve a good balance between source consistency and semantic alignment (closest to the bottom right corner). As a highlight, on the PIE-Bench, our method based on SD-1.5 can get a very close performance to the strong baseline FlowEdit based on SD-3. This further demonstrates our effectiveness. We provide complete quantitative results in the Appendix.
\begin{figure*}[!t]
    \centering
    \includegraphics[width=0.75\linewidth]{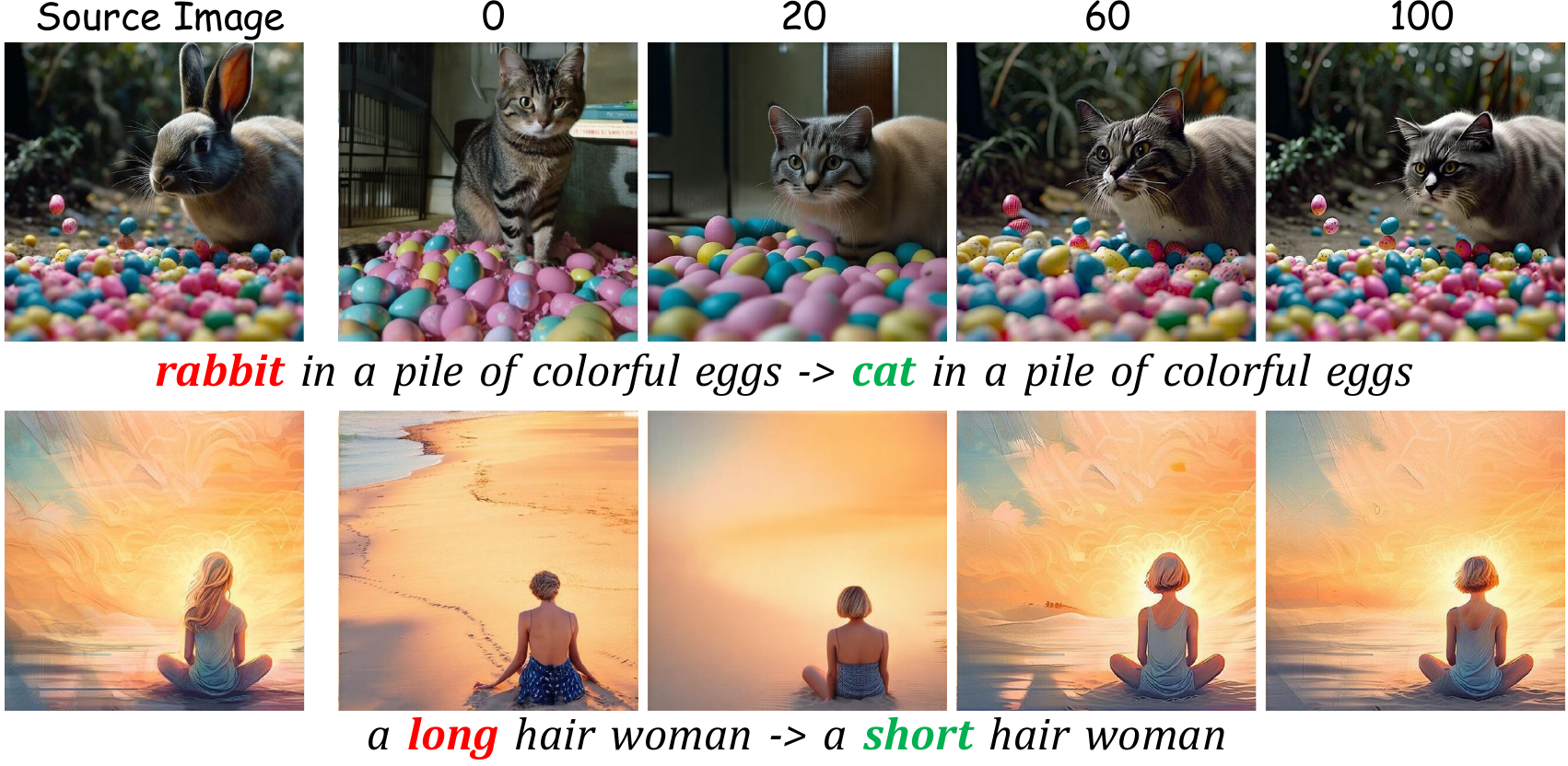}
        \caption{\textbf{Optimization rounds ablation}. From left to right, the quality of the editing is gradually improved (especially for source consistency) across optimization rounds.}
    \label{fig:6}
\end{figure*}

\subsection{Ablation Study}
\label{sec:4.2}

\noindent\textbf{Optimization Rounds}. To analyze how optimization with cycle constraints influences the editing performance, we provided visualized examples in Figure~\ref{fig:6}. Each example contained the editing results of different optimization rounds ($0$,$20$,$60$,$100$). Before the optimization (\ie, adding a random noise), it already achieves a good target semantic alignment. For example, the \texttt{cat} has already appeared to replace the \texttt{rabbit} ($0$ round of the first row), and the \texttt{girl's short hair} has also appeared to replace the \texttt{long hair} ($0$ round of the second row). However, the editing-irrelevant parts (\eg, \texttt{background}, \texttt{candies}, and \texttt{girl}) are quite different from the source images. As the optimization proceeds, the source consistency has gradually improved, while the target semantics remain well-aligned. This indicates that FlowCycle can maintain the editing-irrelevant parts while accomplishing edit the target parts. Quantitative results in the Appendix.

\begin{figure*}[!t]
    \centering
    \includegraphics[width=1\linewidth]{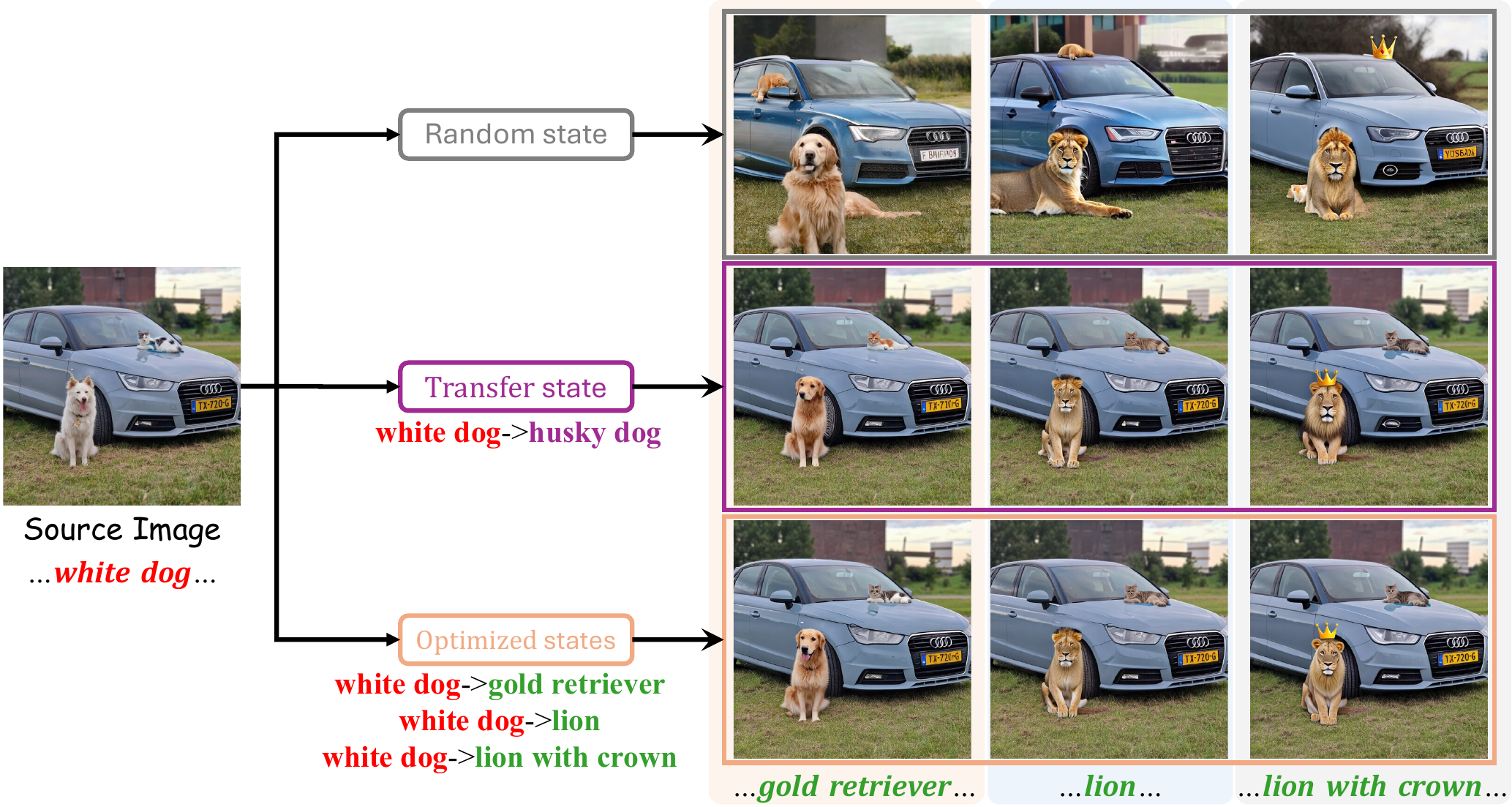}
        \caption{\textbf{Transfer ability of intermediate state}. It shows that the intermediate state (\eg, obtained from ``white dog'' into ``husky dog'' pair) can work well for other targets that have a similar edit pattern (\eg, ``white dog'' into ``golden retriever'').
        }
    \label{fig:7}
\end{figure*}

\noindent\textbf{Intermediate State Transfer}. To further verify our main claim (\ie, the target-aware corruption can make the editing-relevant part editable while maintaining the editing-irrelevant part), we conducted an intermediate state transfer experiment.
As shown in Figure~\ref{fig:7}, given a source image and its source prompt ``\texttt{$\dots$white dog$\dots$}''. We first got three different intermediate states:

\noindent 1) \textit{Random State}: It was acquired by adding a random noise to the source image.

\noindent2) \textit{Transfer State}: We optimized a noise under the ``\texttt{$\dots$white dog$\dots$}'' (source) - ``\texttt{$\dots$husky$\dots$}'' (target) prompt pair using our method. Then we acquired the transfer state by adding this noise to the source image.

\noindent3) \textit{Optimized States}:  We optimized three noises under the subsequent corresponding editing prompt pairs, respectively. Then we acquired three different optimized states by adding these noises to the source image, respectively.

After getting the above states, we edited the source image by following three target prompts, \ie, ``\texttt{$\dots$gold retriever$\dots$}'', ``\texttt{$\dots$lion$\dots$}'', and ``\texttt{$\dots$lion with crown$\dots$}''. As shown in Figure~\ref{fig:7}, we have two observations: 1) The \textit{Transfer State} obviously can lead to better editing results than \textit{Random States}. This demonstrates there is good transferable ability of target-aware intermediate states across similar editing patterns (\eg, change the white dog to another animal). 2) Although the \textit{Transfer State} gets good performance when transferring between cases with similar patterns, it still can not outperform the \textit{Optimized States}. Because the \textit{Optimized States} have more accurate target-aware corruption and source maintaining capacity, leading to the best results among them. This experiment verifies the claim and significance of target-aware editing, and it also provides a protocol to reuse the optimized intermediate states.

\noindent\textbf{Editing Deviation Magnitudes}. As shown in the first row of Figure~\ref{fig:7}, from left to right, the editing target deviation from the source image is increasing (from ``white dog $\rightarrow$ golden retriever'' to ``white dog $\rightarrow$ lion with crown''). We can see that the background preservation is also becoming worse along with the increasing deviation. This demonstrates the limitation of the current target-aware manner, \ie, not flexible to varied editing deviations.
In contrast, our target-aware states (the third row) can adapt better to this type of large-deviation editing (with consistent background preservation).
This indicates that target-aware states can be adjusted according to the specific editing target, thus making it more flexible for varied editing deviation magnitudes.
\begin{table}[!t]
\centering
\caption{Computational overhead comparisons between our accelerated method and other methods on PIE-Bench with SD3. Results are based on a single A800-80GB GPU.}
\resizebox{0.86\columnwidth}{!}
{%
\begin{tabular}{llcccc}
\hline\hline
                & Method   & Distance $\downarrow$ & CLIP $\uparrow$ & Time (s) & GPU (GB)  \\ \hline
\multirow{2}{*}{\begin{tabular}[c]{@{}l@{}}Optimization-free\end{tabular}}             & FlowAlign & 0.028      & 25.28 & 4 & 16\\
           & Qwen-Image-Edit  & 0.054      & 25.20 & 92 & 60\\ \hline
\multirow{5}{*}{\begin{tabular}[c]{@{}l@{}}Optimization-based\end{tabular}} 
                    & iRFDS & 0.069 & 25.12 & 132 & 34\\
                      & \cellcolor{mygray}Ours       & \cellcolor{mygray}0.013      & \cellcolor{mygray}25.48 & \cellcolor{mygray}247 & \cellcolor{mygray}16\\
                      & \cellcolor{mygray}Ours-accelerate-weak         & \cellcolor{mygray}0.014      & \cellcolor{mygray}25.41 & \cellcolor{mygray}160 & \cellcolor{mygray}16\\ 
                      & \cellcolor{mygray}Ours-accelerate-mid         & \cellcolor{mygray}0.014      & \cellcolor{mygray}25.34 & \cellcolor{mygray}126 & \cellcolor{mygray}16\\ 
                      & \cellcolor{mygray}Ours-accelerate-strong         & \cellcolor{mygray}0.017      & \cellcolor{mygray}25.31 & \cellcolor{mygray}90 & \cellcolor{mygray}16\\ \hline

                      \hline\hline
\end{tabular}%
}
\label{tab:1}
\end{table}

\subsection{Computational Overhead Analysis}
\label{sec:4.3}

We gave computational overhead (including editing time per image and required GPU memory) comparisons between the current optimization-free methods, the optimization-based methods, and our FlowCycle.

According to Section~\ref{sec:3.4}, we can accelerate our optimization process for faster editing. To this end, we gave the results of three different acceleration versions. 1) \textit{Ours-accelerate-weak}: the $f(n)=\text{rounding}(sin\frac{\pi n}{2N}*M)$ and it can accelerate around $1.5$ times over original FlowCycle. 2) \textit{Ours-accelerate-mid}: the $f(n)=\text{rounding}(\frac{nM}{N})$, and it can accelerate around two times over the original FlowCycle. 3) \textit{Ours-accelerate-strong}: the $f(n)=\text{rounding}((1-con\frac{\pi n}{2N})*M)$, and it can accelerate around three times over the original FlowCycle.

As shown in Table~\ref{tab:1}, we can have three observations: 1) The acceleration operation can significantly reduce the editing time (\eg, the time overhead of the ``strong'' version is only around $1/3$ of the no-acceleration version) with a little performance degradation. 2) Compared with the SOTA optimization-free method \textit{FlowAlign}, our method can gain better editing quality at the cost of a higher time cost. 3) Compared with the large editing model based method \textit{Qwen-Image} and optimization-based \textit{iRFDS}, \textit{Ours-accelerate-strong} can outperform them on both editing quality and cost. 

\section{Conclusion}
\label{sec:5}
In this paper, we revealed the limitations of the target-agnostic intermediate state in the existing corruption-then-restoration editing paradigm and proposed to acquire target-aware intermediate states by optimizing two noises with cycle-consistent constraints. To satisfy consistency, the editing result is improved to maintain the editing-irrelevant contents of the source image for better source consistency. We conducted extensive experiments to show the effectiveness and significance of the target-aware manner. 
%
%
\bibliographystyle{splncs04}
\bibliography{main}

\end{document}